\documentclass{article} 
\usepackage{iclr2027_conference,times}

\usepackage{amsmath,amsfonts,bm}

\def\eqref#1{equation~\ref{#1}}

\def\1{\bm{1}}

\DeclareMathAlphabet{\mathsfit}{\encodingdefault}{\sfdefault}{m}{sl}
\SetMathAlphabet{\mathsfit}{bold}{\encodingdefault}{\sfdefault}{bx}{n}

\usepackage[utf8]{inputenc} 
\usepackage[T1]{fontenc}    
\usepackage{hyperref}       
\usepackage{url}            
\usepackage{booktabs}       
\usepackage{amsfonts}       
\usepackage{nicefrac}       
\usepackage{microtype}      
\usepackage{xcolor}         

\usepackage{graphicx}
\usepackage{amsmath}
\usepackage{multirow}
\usepackage{amssymb}
\usepackage{wrapfig}

\title{Structured Residual Connectivity Matters for Diffusion Transformers}

\author{
Yuhe Liu$^{1}$ \quad Xinyin Ma$^{1}$ \quad Gongfan Fang$^{1}$ \quad Songhua Liu$^{2}$ \quad Xinchao Wang$^{1}$\thanks{Corresponding Author.}\\
$^{1}$National University of Singapore \quad 
$^{2}$Shanghai Jiao Tong University\\
{\tt\small \{liuyuhe, maxinyin, gongfan\}@u.nus.edu \quad liusonghua@sjtu.edu.cn \quad xinchao@nus.edu.sg}
}

\iclrfinalcopy 
\begin{document}

\maketitle
\lhead{}\renewcommand{\headrulewidth}{0pt} 

\vspace{-3mm}
\begin{abstract}

Diffusion Transformers (DiTs) have established themselves as a scalable backbone for high-fidelity image synthesis. However, unlike U-Net based diffusion models that rely on rigid, hand-crafted skip connections, DiTs predominantly use a uniform residual stream that integrates all preceding layers as a monolithic state. 
In this work, we rethink residual connections in diffusion transformers and propose to transform them from passive summation into an active retrieval mechanism optimized for image denoising.
First, we conduct a systematic analysis of DiT's internal representation, revealing a latent preference for early-layer feature reuse and symmetric layer guidance. Motivated by this, we introduce a structured connectivity design that explicitly integrates local residual connections with long-range pathways.
Instead of static skip connections or dense all-layer routing, our method enables each transformer block to selectively ``attend'' to critical earlier representations, dynamically retrieving spatial and semantic cues through direct, differentiable cross-depth paths.
Experiments show that our adaptive connectivity leads to faster convergence, with up to $1.73\times$ fewer training iterations, and significant gains in FID and visual quality with less than $0.1\%$ additional parameters, further improving a strong REPA-XL/2 model from $5.9$ to $4.34$ FID without guidance and reaching $1.39$ FID with classifier-free guidance. Our findings suggest that adaptive cross-layer connectivity is a critical yet underexplored factor in diffusion transformers, and that incorporating structured information pathways provides a simple and effective direction for improving scalable generative models.

\end{abstract}

\section{Introduction}

The design of residual pathways has been a central problem in deep neural networks. Residual connections~\citep{he2016deep} enable stable optimization by allowing features to bypass non-linear transformations, while subsequent architectures such as Highway Networks~\citep{srivastava2015highway} and DenseNet~\citep{huang2017densely} further enhance information flow through gated or dense connectivity. These designs demonstrate that how information is propagated across layers plays a crucial role in optimization, representation learning, and scalability. More recently, similar trends have been observed in large language models (LLMs), where architectural modifications such as Hyper-Connections, mHC, and Attention Residuals~\citep{zhu2024hyper, xie2025mhc,chen2026attnres} are introduced to improve information flow and mitigate gradient vanishing and representation collapse, further highlighting the importance of residual connections in model architectures.

In generative modeling, this structural design is a defining factor for performance. Traditional UNet-based architectures~\citep{ronneberger2015u} thrive on long-range skip connections, which provide a strong inductive bias by explicitly preserving multi-scale features. In contrast, Diffusion Transformers (DiTs)~\citep{peebles2023scalable} represent a paradigm shift toward scalability, utilizing a uniform residual stream to manage information flow. This shift introduces a fundamental discrepancy in residual design: UNets maintain distinct, structured pathways for different levels of abstraction, while DiTs rely on local additive updates that conflate all preceding layers into an additive hidden state. This uniform accumulation progressively dilutes the contribution of individual layers and, more critically, imposes a rigid topology on the backward pass, limiting the model's ability to adaptively optimize gradient flow across different denoising stages. 


\begin{figure}
    \centering
    \includegraphics[width=0.93\linewidth]{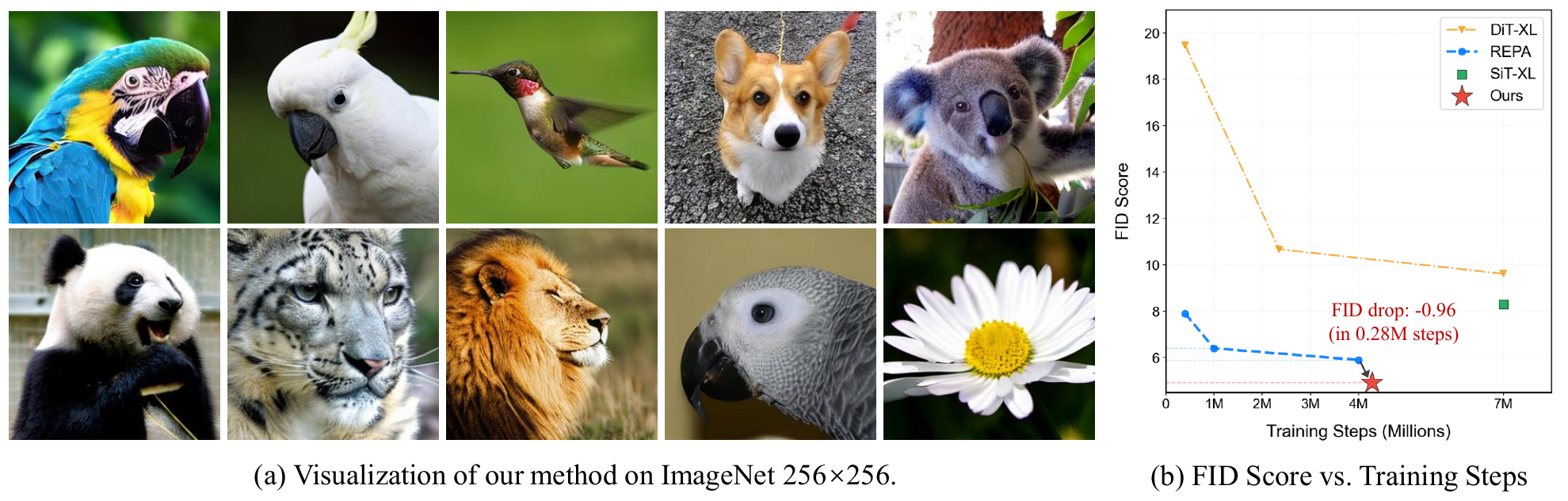}
\vspace{-2mm}
    \caption{
Qualitative and quantitative results of our method on ImageNet $256\times256$.
(a) Selected generated samples from our model on REPA-XL/2 fine-tuning. We use
classifier-free guidance with $w = 4.0$.
(b) FID comparison under different training budgets. With only $0.28$M additional training steps, our method further improves a strong REPA-XL/2 model from $5.9$ to $4.94$ FID, 
whereas the continued fine-tuning baseline with the same budget remains at $5.87$ (Table~\ref{tab:main}), 
demonstrating that structured residual routing can effectively enhance an already well-trained model. 
}
\vspace{-5mm}
    \label{fig:intro}
\end{figure}


This leads to a key question: how should residual pathways be designed for diffusion transformers? In particular, can we combine the scalability of transformer architectures with the structured connectivity patterns that enable effective generative modeling?

Prior works have explored this direction from two sides. U-ViT~\citep{bao2023all} and U-DiTs~\citep{tian2024u} bring UNet-style long skip connections back into transformer backbones, but merge the skipped features with a static learned projection. Attention Residuals~\citep{chen2026attnres} in LLMs make residual aggregation input-dependent, but route densely over all preceding layers.

In this work, we revisit residual pathways in DiTs from the perspective of information flow. We find that the standard residual stream homogenizes representations across depth (Figure~\ref{fig:analysis}(a)), whereas, when granted the freedom to route across layers, DiTs consistently rely on early-layer representations and exhibit a preference for mirrored/symmetric layer pairs (Section~\ref{sec:allrouting}). Driven by these findings, we propose a Structured Connectivity Design that transforms the residual stream from passive summation into active retrieval along structured cross-layer paths, allowing gradients to flow directly back to stage-matched encoder layers.
With less than $0.1\%$ additional parameters, our method consistently improves DiTs across model scales and reaches the baseline FID with up to $1.73\times$ fewer iterations. As shown in Figure~\ref{fig:intro}(b), it further improves a strong REPA-XL/2 model from $5.9$ to $4.94$ FID with only $0.28$M additional iterations and further to $4.34$ FID at $0.35$M steps (Table~\ref{tab:main}), achieving significantly better performance and accelerating the convergence of REPA. With classifier-free guidance, our model further reaches $1.39$ FID (Table~\ref{tab:main_cfg}).

Our results highlight that adaptive cross-layer connectivity is a critical yet underexplored factor in diffusion transformers, and that reintroducing structured information pathways provides a simple and effective direction for improving scalable generative models.

\section{Related work}

\paragraph{Generative models.}
Generative modeling has been a central topic in machine learning, with different paradigms developed to approximate complex data distributions.
Deep generative models include variational autoencoders~\citep{kingma2013auto}, generative adversarial networks~\citep{goodfellow2014generative}, and autoregressive models~\citep{chen2020generative, li2024autoregressive, tian2024visual}, while diffusion models~\citep{ho2020denoising, song2020score} have recently become a dominant framework for high-fidelity image generation.
Subsequent works improve diffusion models from different perspectives, including better likelihood and faster sampling~\citep{nichol2021improved, zhou2024simple, salimans2022progressive, zhou2025few}, non-Markovian sampling processes~\citep{song2020denoising, chen2024fast}, representation alignment~\citep{yu2024representation}, and architectural design~\citep{xie2024sana, peebles2023scalable, ma2024sit, bao2023all, tian2024u}.
Guided diffusion further shows that architectural and guidance choices can significantly improve generation quality~\citep{dhariwal2021diffusion, tan2025ominicontrol, liu2026gated}.

\paragraph{Model structure design.}
Model structure design is crucial for optimization, representation learning, and scalability.
Residual connections~\citep{he2016deep} enable stable training through identity shortcuts, while Highway Networks~\citep{srivastava2015highway} and DenseNet~\citep{huang2017densely} further improve information flow through gated or dense cross-layer pathways.
Recent works in large language models also revisit residual pathways to improve information propagation and mitigate issues such as attention collapse or representation collapse~\citep{qiu2025gated, zhu2024hyper, xie2025mhc, zhang2026xhc, chen2026attnres}.
In generative vision models, UNet architectures~\citep{ronneberger2015u} rely on hierarchical encoder-decoder structures and skip connections to preserve spatial details and reuse multi-level features.
Diffusion Transformers (DiTs)~\citep{peebles2023scalable} replace UNet backbones with scalable transformer blocks operating on latent patches; follow-up works such as U-ViT~\citep{bao2023all} and U-DiTs~\citep{tian2024u} reintroduce long skip connections with static concatenation-and-projection fusion, and DDT~\citep{wang2026ddt} decouples the model into a condition encoder and a velocity decoder conditioned on the final encoder output.
In contrast, guided by a structural analysis of routing in DiTs, our method lets each decoder layer dynamically draw on its stage-matched encoder representation with content-dependent weights.

\begin{figure}[t]
    \centering
    \includegraphics[width=0.78\linewidth]{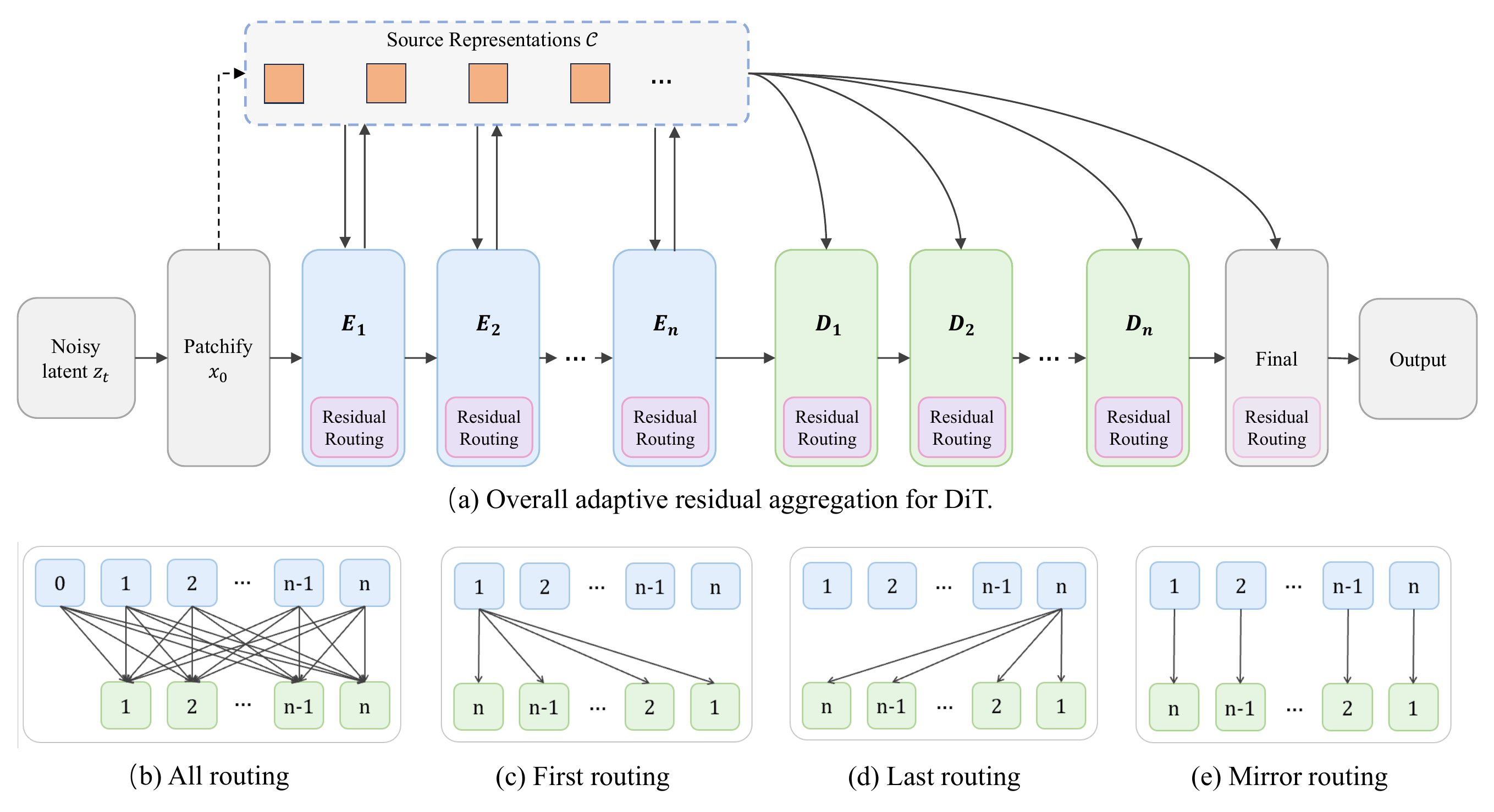}
    \vspace{-2mm}
    \caption{
    Overview of the proposed adaptive residual aggregation for DiT.
    (a) Encoder-side representations are collected as differentiable source representations $\mathcal{C}_{\mathrm{enc}}$ and routed to later layers for residual reconstruction.
    (b)--(e) illustrate different source routing strategies, including all, first, last, and mirror routing.
    Our final design adopts mirror routing, which provides a sparse and stage-matched source path for decoder-side residual reconstruction.
    }
    \label{fig:model1}
\vspace{-5mm}
\end{figure}

\section{Method}

\subsection{Preliminary: residuals as depth-wise routing}

Diffusion Transformers (DiTs) typically adopt standard residual connections inside each transformer block.
Let $x_l \in \mathbb{R}^{N \times d}$ denote the token representation before layer $l$, where $N$ is the number of latent patches and $d$ is the hidden dimension.
A standard residual update can be written as:
\begin{equation}
x_{l+1} = x_l + f_l(x_l),
\end{equation}
where $f_l(\cdot)$ denotes the transformation at layer $l$, such as self-attention or MLP.
Unrolling this recurrence shows that the final representation implicitly accumulates previous layer outputs with fixed unit coefficients:
\begin{equation}
x_L = x_0 + \sum_{l=0}^{L-1} f_l(x_l).
\end{equation}
Therefore, residual connections are not only optimization shortcuts, but also define how information is routed and aggregated across depth.
However, standard residual connections use \textbf{fixed} and \textbf{uniform} aggregation, without a mechanism to selectively emphasize useful intermediate representations.

A more flexible alternative is attention-based residual routing, where each layer adaptively aggregates historical representations through learnable weights, as shown in~\citet{chen2026attnres}.
Nevertheless, dense all-to-all routing must retain all preceding representations as candidate sources, which increases memory and computation and may lead to redundant feature mixing.
In this work, we preserve adaptive residual aggregation while replacing \textbf{dense} routing with a \textbf{sparse} and \textbf{structured} path.

\subsection{Overview: differentiable residual collection and routing}

\begin{wrapfigure}{r}{0.44\linewidth}
    \centering
    \vspace{-8pt}
    \includegraphics[width=0.9\linewidth]{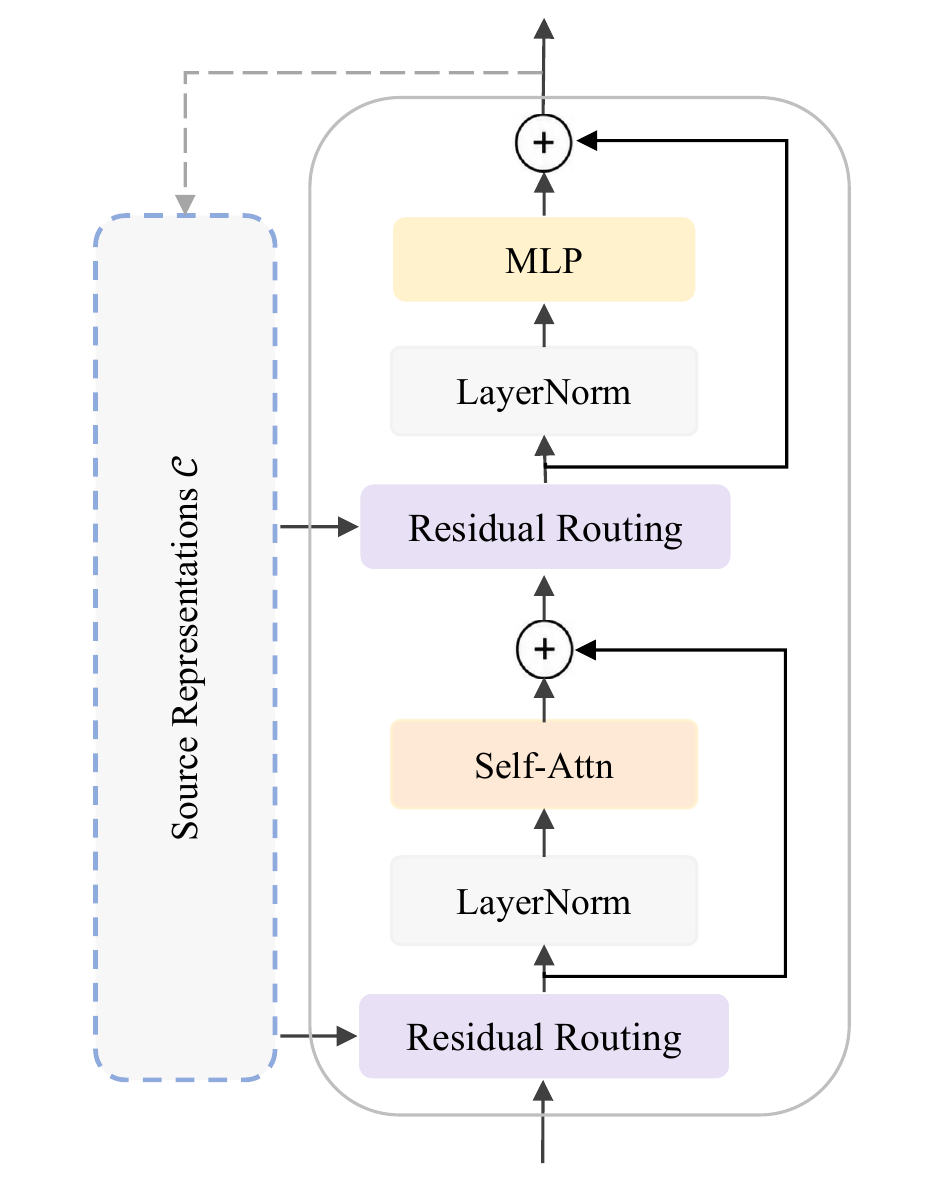}
    \caption{
    Residual routing inside a DiT block.
    The operator reconstructs each sublayer input from the current representation and selected source representations $\mathcal{C}$.
    }
    \label{fig:model2}
    \vspace{-5pt}
\end{wrapfigure}

Figure~\ref{fig:model1} illustrates the overall framework.
Given the noisy latent $z_t$, we first obtain the initial token representation $x_0$ through patch and positional embedding.
In the standard DiT architecture, all transformer blocks operate at the same token resolution and follow a homogeneous single-scale design, which contributes to its \textbf{generality} and \textbf{scalability}.
Nevertheless, prior works~\citep{wang2026ddt, tumanyan2023plug} suggest that even in such homogeneous architectures, different depths may play different functional roles during generation.
As depth increases, the model tends to progressively abstract semantic representations from fine-grained features, and then leverage these semantic representations to guide the refinement and reconstruction of visual details.
Motivated by this perspective, we evenly divide DiT along the depth dimension into an encoder phase and a decoder phase, while keeping its original single-scale token resolution unchanged.

\paragraph{Residual collection.}
We collect intermediate representations as differentiable residual sources:
\begin{equation}
\mathcal{C}_{\mathrm{enc}} = \{x_0, x_1, x_2, \ldots, x_K\},
\end{equation}
where $x_i$ denotes the output of the $i$-th encoder block or decoder block. The patch embedding output $x_0$ before the first encoder block is inserted into the encoder memory.
Here, the output of patch embedding is also included as a residual source, providing direct access to the patch-level representation before any transformer block.
These sources are not detached from the computation graph, so gradients from decoder-side routing flow back to the encoder-side representations. This distinguishes our sources from a static feature cache and enables direct cross-depth optimization.

\subsection{Residual routing operator}
\label{sec:operator}


We now define how a selected source set is used inside each DiT block.
Given source representations $\mathcal{C}=\{c_1,\ldots,c_M\}$ and an optional current representation $x$, we define the candidate set:
\begin{equation}
\mathcal{V} =
\begin{cases}
\mathcal{C} \cup \{x\}, & x \neq \varnothing, \\
\mathcal{C}, & x = \varnothing.
\end{cases}
\end{equation}
Then 
we compute routing weights over $\mathcal{V}$ via a softmax selection mechanism.
For each candidate $v_i \in \mathcal{V}$, we first normalize it and compute its routing logit and weight:
\begin{equation}
s_i = w^\top \mathrm{RMSNorm}(v_i), \qquad
\alpha_i = \frac{\exp(s_i)}{\sum_{j=1}^{|\mathcal{V}|}\exp(s_j)},
\end{equation}
where $w \in \mathbb{R}^{d}$ is a learnable routing vector.
The routed representation is then obtained by:
\begin{equation}
\mathcal{R}(\mathcal{C},x) = \sum_{i=1}^{|\mathcal{V}|}\alpha_i v_i .
\end{equation}
The operator $\mathcal{R}(\mathcal{C},x)$ adaptively reconstructs the residual source from the current representation and the selected source representations.
For a DiT sublayer with input $x_l$ and source set $\mathcal{C}_l$, we first compute the routed representation and then apply the sublayer transformation to it:
\begin{equation}
\tilde{x}_l = \mathcal{R}(\mathcal{C}_l,x_l), \qquad
x_{l+1} = \tilde{x}_l + f_l(\tilde{x}_l),
\end{equation}
where $f_l(\cdot)$ denotes the forward pass of the corresponding sublayer.
As shown in Figure~\ref{fig:model2}, we insert this operator before both the self-attention and MLP sublayers.
Therefore, the residual source is no longer restricted to the identity stream, but can selectively incorporate useful features from the selected source representations.

\subsection{Routing on all layers}
\label{sec:allrouting}
\begin{figure}[t]
    \centering
    \includegraphics[width=0.95\linewidth]{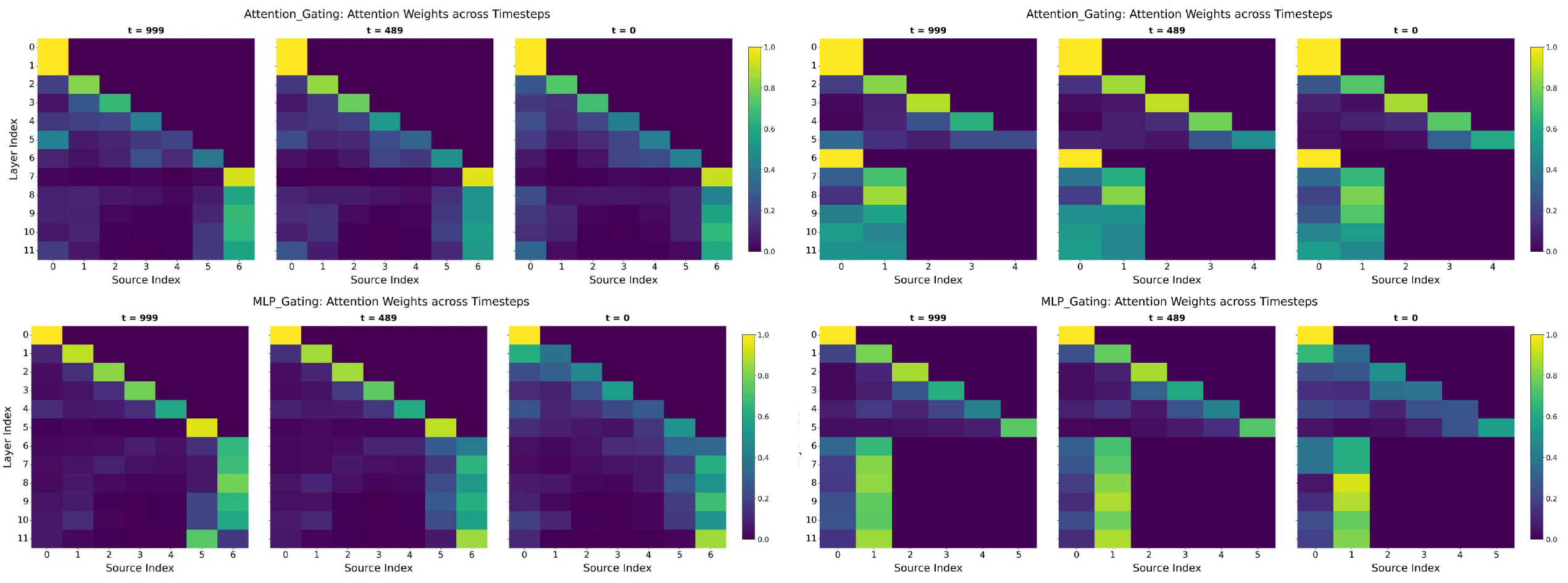}
    \vspace{-2mm}
    \caption{
    Attention weights across layers under different connectivity designs and diffusion timesteps. 
    Each row corresponds to a target layer, and each column denotes the source layers being attended to.
    (Left) Attention-based residual across all layers, which exhibits a preference for mirrored/symmetric layer pairs. 
    (Right) With symmetric (mirror) connections, attention concentrates on corresponding layers, indicating a strong preference for structured connectivity.
    }
    \label{fig:attention}
\vspace{-5mm}
\end{figure}

We first conduct a preliminary experiment to use the representations of all encoder layers as the source representations $\mathcal{C}$, replacing static residuals with the learned routed representations. We visualize the routing weight $\alpha_i$ for each layer, and the results are shown in Figure~\ref{fig:attention}. Our observations yield two insights that deviate from standard transformer behavior:
\begin{itemize}
    \item Encoder-Side Dependency: Contrary to the local-dominance patterns often seen in language models, DiTs consistently assign high importance to early-layer representations across the entire depth of the network. This suggests that ``encoder-side'' spatial cues are indispensable for maintaining structural integrity during denoising.  
    \item Spontaneous Symmetry Bias: Most notably, when granted the structural freedom to attend across layers, the model exhibits a preference for mirrored/symmetric layer pairs. This behavior suggests that symmetric pathways are not merely a heuristic design of UNets, but a latent structural necessity that the model seeks out to stabilize its gradient flow.  
\end{itemize}

\subsection{Source routing strategies}
\label{sec:strategies}

Motivated by these observations, we further design how decoder layers reuse the collected representations through a routing strategy.
As shown in Figure~\ref{fig:model1}(b)--(e), different choices of the source subset lead to different residual routing patterns.
Dense all routing allows each decoder layer to access all collected sources, but it can be computationally redundant and may introduce ambiguous feature routing.
Our visualization in Figure~\ref{fig:attention} also shows that the first and last encoder-side representations receive relatively high weights, motivating us to include first routing and last routing as comparison settings.
However, such fixed single-source strategies lack stage-wise correspondence.
In contrast, mirror routing assigns each decoder layer to its corresponding encoder-side source, yielding a structured and generalizable design through sparse stage-matched routing, which empirically achieves the best FID among these strategies.

Given the encoder-side sources $\mathcal{C}_{\mathrm{enc}}$, each decoder layer selects a source subset for residual routing.
The decoder is initialized by the last encoder representation $x_K$.
For decoder layer $f_{K+j}$, where $j \in \{1,\ldots,K\}$, we select a source subset $\mathcal{C}^{\mathrm{dec}}_j \subseteq \mathcal{C}_{\mathrm{enc}}$ and update:
\begin{equation}
x_{K+j+1} = f_{K+j}(x_{K+j}, \mathcal{C}^{\mathrm{dec}}_j),
\end{equation}
where $x_{K+j+1}$ denotes the output of the $j$-th decoder layer. We consider the following strategies.

\paragraph{All routing.}
Each decoder layer can access the full encoder-side source set:
\begin{equation}
\mathcal{C}^{\mathrm{all}}_j = \mathcal{C}_{\mathrm{enc}}.
\end{equation}

\paragraph{First/Last routing.}
Each decoder layer only accesses the first or the last encoder-side representation after the initial patch-level source:
\begin{equation}
\mathcal{C}^{\mathrm{first}}_j = \{x_1\} \quad \text{or} \quad \mathcal{C}^{\mathrm{last}}_j = \{x_K\}.
\end{equation}

\paragraph{Mirror routing.}
For our final design, each decoder layer accesses the source representation from its mirrored encoder stage, with mirror index $m(j)=K-j+1$:
\begin{equation}
\mathcal{C}^{\mathrm{mirror}}_j = \{x_{m(j)}\}.
\end{equation}

Compared with all routing, mirror routing replaces dense access to all encoder-side sources with a single structured source for each decoder layer.
It therefore reduces routing ambiguity while preserving sparse, stage-matched cross-depth interaction.

\subsection{Discussion}
\label{sec:discussion}

Our method differs from Attention Residuals~\citep{chen2026attnres} for LLMs in two aspects of routing structure, despite sharing the softmax source-scoring mechanism.
First, we restrict the source pool to encoder-side representations rather than all preceding layer outputs.
Second, each decoder layer retrieves a single stage-matched mirror source instead of routing densely over the source pool.
These choices tailor residual routing to image generation, motivated by the encoder-side dependency and symmetry bias observed in Section~\ref{sec:allrouting}.
Meanwhile, Hyper-Connections and their latest variants mHC and xHC~\citep{zhu2024hyper, xie2025mhc, zhang2026xhc} in LLMs address a different design dimension: they expand or mix parallel residual streams around each layer without explicitly selecting earlier-layer representations as cross-depth sources.

Mirror routing resembles the long skip connections of U-ViT~\citep{bao2023all} and U-DiTs~\citep{tian2024u} in stage pairing, but differs in how the paired features are fused.
These previous methods use learned but static projections whose fusion weights do not adapt to individual tokens or denoising stages.
Our routing weights instead depend on the token representations and adapt across inputs, spatial positions, sublayers, and diffusion timesteps.
Our design thus enables adaptive feature reuse along structured cross-layer paths while preserving DiT's single-scale architecture.

\section{Experiments}

\subsection{Experimental setup}

\paragraph{Dataset and evaluation.}
We evaluate class-conditional ImageNet $256\times256$ generation in the latent space following standard latent diffusion protocols, and report FID, sFID, Inception Score (IS), Precision, and Recall; unless otherwise specified, results are reported without classifier-free guidance.

\paragraph{Models.}
We evaluate our method on two families of diffusion transformers with different parameterizations:
DiT-S/2, DiT-B/2, and DiT-XL/2~\citep{peebles2023scalable}, trained from scratch with $\epsilon$-prediction under the DDPM formulation,
and SiT-XL/2~\citep{ma2024sit} + REPA~\citep{yu2024representation}, trained with velocity prediction under a linear interpolant and fine-tuned from its released 4M-iteration checkpoint.
Our routing
adds only lightweight normalization and projection layers.

\paragraph{Baselines and variants.}
Besides standard DiTs, we compare with
(i) our DiT implementation of Attention Residuals (AttnRes)~\citep{chen2026attnres}, which densely routes over all preceding block outputs;
(ii) a U-ViT-style static mirror skip~\citep{bao2023all}, which connects the same mirror pairs via concatenation followed by a learned linear projection;
(iii) Fused-Mirror without encoder interaction; and
(iv) fused residual routing with all, first, last, or mirror source selection (Fused-All/First/Last/Mirror).
All variants use the same training protocol, which allows us to disentangle the effects of residual formulation and connectivity.
\subsection{Main results}
\label{sec:main}

\begin{table*}[t]
\centering
\begin{minipage}[t]{0.425\linewidth}
\centering
\caption{
\textbf{FID comparisons with DiTs and SiTs} on ImageNet $256\times256$ without CFG.
DiT baselines are reproduced under our training protocol.
Cont.\ FT: continued fine-tuning of the released REPA checkpoint with the same additional budget.
Our routing adds $<$0.1\% parameters.
}
\label{tab:main}
\vspace{0mm}
\setlength{\tabcolsep}{3pt}
\renewcommand{\arraystretch}{1.0}
\scriptsize
\begin{tabular}{lccc}
\toprule
Model & \#Params & Iter. & FID$\downarrow$ \\
\midrule
DiT-S/2 & 33M & 400K & 69.81 \\
\textbf{+ Ours} & 33M+0.019M & \textbf{400K} & \textbf{62.48} \\
\midrule
DiT-B/2 & 130M & 400K & 44.21 \\
\textbf{+ Ours} & 130M+0.038M & \textbf{400K} & \textbf{36.33} \\
\midrule
DiT-XL/2 & 675M & 400K & 18.85 \\
\textbf{+ Ours} & 675M+0.13M & \textbf{400K} & \textbf{15.07} \\
\midrule
DiT-XL/2 & 675M & 800K & 13.71 \\
DiT-XL/2 & 675M & 1300K & 11.76 \\
\textbf{+ Ours} & 675M+0.13M & \textbf{800K} & \textbf{11.40} \\
\midrule
SiT-XL/2 & 675M & 7M & 8.3 \\
+ REPA & 675M & 4M & 5.9 \\
+ REPA, Cont.\ FT & 675M & 4M+0.28M & 5.87 \\
\textbf{+ REPA + Ours} & 675M+0.13M & \textbf{4M+0.28M} & \textbf{4.94} \\
+ REPA, Cont.\ FT & 675M & 4M+0.35M & 6.01 \\
\textbf{+ REPA + Ours} & 675M+0.13M & \textbf{4M+0.35M} & \textbf{4.34} \\
\bottomrule
\end{tabular}
\end{minipage}
\hfill
\begin{minipage}[t]{0.565\linewidth}
\centering
\caption{
\textbf{System-level comparison} on ImageNet $256\times256$ with CFG.
Epochs reported by prior works are converted to iterations assuming $\sim$5K iterations per epoch.
Results with additional CFG scheduling are marked with an asterisk ($^*$); REPA and ours use the same guidance interval $[0, 0.7]$.
}
\label{tab:main_cfg}
\vspace{0mm}
\setlength{\tabcolsep}{3pt}
\renewcommand{\arraystretch}{0.92}
\scriptsize
\begin{tabular}{lcccccc}
\toprule
Model & Iter. & FID$\downarrow$ & sFID$\downarrow$ & IS$\uparrow$ & Pre.$\uparrow$ & Rec.$\uparrow$ \\
\midrule
\multicolumn{7}{l}{\textit{Pixel diffusion}} \\
\quad ADM-U & 2M & 3.94 & 6.14 & 215.8 & 0.83 & 0.53 \\
\quad VDM++ & 2.8M & 2.40 & -- & 225.3 & -- & -- \\
\quad Simple diffusion & 4M & 2.77 & -- & 211.8 & -- & -- \\
\quad CDM & 10.8M & 4.88 & -- & 158.7 & -- & -- \\
\midrule
\multicolumn{7}{l}{\textit{Latent diffusion, U-Net}} \\
\quad LDM-4 & 1M & 3.60 & -- & 247.7 & \textbf{0.87} & 0.48 \\
\midrule
\multicolumn{7}{l}{\textit{Latent diffusion, Transformer + U-Net hybrid}} \\
\quad U-ViT-H/2 & 1.2M & 2.29 & 5.68 & 263.9 & 0.82 & 0.57 \\
\quad DiffiT$^*$ & -- & 1.73 & -- & 276.5 & 0.80 & 0.62 \\
\quad MDTv2-XL/2$^*$ & 5.4M & 1.58 & 4.52 & \textbf{314.7} & 0.79 & \textbf{0.65} \\
\midrule
\multicolumn{7}{l}{\textit{Latent diffusion, Transformer}} \\
\quad MaskDiT & 8M & 2.28 & 5.67 & 276.6 & 0.80 & 0.61 \\
\quad SD-DiT & 2.4M & 3.23 & -- & -- & -- & -- \\
\quad DiT-XL/2 & 7M & 2.27 & 4.60 & 278.2 & 0.83 & 0.57 \\
\quad SiT-XL/2 & 7M & 2.06 & 4.50 & 270.3 & 0.82 & 0.59 \\
\qquad + REPA & 1M & 1.96 & \textbf{4.49} & 264.0 & 0.82 & 0.60 \\
\qquad + REPA$^*$ & 4M & 1.42 & 4.70 & 305.7 & 0.80 & \textbf{0.65} \\
\qquad \textbf{+ REPA + Ours}$^*$ & \textbf{4M+0.28M} & \textbf{1.39} & \textbf{4.49} & 297.4 & 0.79 & 0.64 \\
\bottomrule
\end{tabular}
\end{minipage}
\vspace{-4mm}
\end{table*}

\paragraph{Comparison with DiTs.}
As shown in Table~\ref{tab:main}, under an identical training protocol our routing consistently improves DiTs~\citep{peebles2023scalable} at $400$K iterations, reducing FID from $69.81$ to $62.48$ on DiT-S/2, from $44.21$ to $36.33$ on DiT-B/2, and from $18.85$ to $15.07$ on DiT-XL/2.
The relative FID reduction grows with model size ($10.5\%$, $17.8\%$, and $20.1\%$), while the added parameters remain below $0.1\%$.
With longer training, our method retains this advantage on DiT-XL/2 (Figure~\ref{fig:analysis}(d)).
For example, our model at $800$K iterations ($11.40$ FID) already outperforms the baseline trained for $1300$K iterations ($11.76$).
Across matched-FID comparisons, our method requires up to $1.73\times$ fewer training steps, and this advantage tends to grow as training progresses.

\paragraph{Improving a strong pretrained model.}
We further evaluate whether our method remains effective when applied to a strong representation-enhanced model, SiT-XL/2~\citep{ma2024sit} + REPA~\citep{yu2024representation}.
As shown in Figure~\ref{fig:intro}(b), REPA converges rapidly in the early stage, but its late-stage improvement becomes much slower: after reaching $6.4$ FID at $1$M iterations, it requires another $3$M iterations to improve to $5.9$, yielding only a $0.5$ FID gain.
This suggests that the model is close to saturation under its original architecture and optimization pathway, and matched continued fine-tuning from the released $4$M checkpoint indeed brings no meaningful gain.

At the same time, after adding our routing and fine-tuning for only $0.28$M iterations, the model improves from $5.9$ to $4.94$ FID, a much larger $0.96$ FID reduction with far fewer updates.
Extending fine-tuning to $0.35$M iterations further reduces FID to $4.34$, increasing the total reduction to $1.56$ FID.
This result indicates that the late-stage bottleneck is \textbf{not simply due to insufficient model capacity}. Instead, there remains substantial room for improving the model's connectivity and gradient structure.

These results suggest that structured residual routing can unlock additional expressive capacity from an already strong pretrained model.
Importantly, the benefit is not limited to from-scratch DiT training.
Even when applied as a fine-tuning module on top of a model trained with a different representation-enhancement strategy, our method still provides clear improvements.
With the guidance interval~\citep{kynkaanniemi2024applying}, the $0.28$M model further reaches $1.39$ FID, outperforming REPA ($1.42$) and the Transformer + U-Net hybrids in Table~\ref{tab:main_cfg}.
This demonstrates that \textbf{optimizing residual connectivity and gradient propagation} is broadly beneficial for exploiting the representational potential of diffusion transformers.


\paragraph{Efficiency.}
We analyze the efficiency of our method on DiT-S/2 in Table~\ref{tab:cost} (Appendix~\ref{sec:app_results}).
Our routing is parameter-efficient, adding only $0.019$M parameters and $0.3\%$ FLOPs.
Compared with dense AttnRes under the same parameter budget, it matches the FID ($62.48$ vs.\ $62.87$) while routing each decoder sublayer over a single stage-matched source instead of $O(L)$ sources, which reduces the decoder-side source-stack activation by about $81\%$ ($1.51$ GB $\rightarrow$ $289$ MB).
This saving grows with depth, since dense routing stacks more candidate sources in deeper models.

Overall, the results indicate that improving residual connectivity provides consistent benefits across model scales and training regimes.
The gains are especially meaningful because the parameter overhead is negligible, suggesting that the improvement primarily comes from better information routing rather than increased model capacity.

\subsection{Analysis: representation and gradient structure}

\begin{figure}[t]
    \centering
    \includegraphics[width=0.94\linewidth]{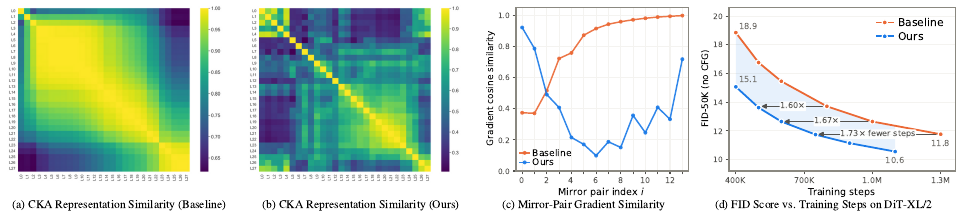}
    \vspace{-2mm}
    \caption{
    Analysis of representation and gradient structure on DiT-XL/2.
    (a) The baseline exhibits highly similar representations across layers, indicating strong feature homogenization.
    (b) Our method reduces excessive feature similarity and introduces clearer depth-wise structure, with higher similarity between the earliest and latest layers.
    (c) Gradient similarity between mirror layer pairs. The $i$-th point measures the cosine similarity between gradients of layer $i$ and layer $27-i$ in DiT-XL/2.
    Our method increases gradient alignment for early--late mirror pairs while preserving differentiation in middle layers.
    (d) FID-50K (w/o CFG) from $400$K to $1.3$M steps. Our method achieves lower FID throughout training and matches the baseline's FID with up to $1.73\times$ fewer steps.
    }
    \label{fig:analysis}
\vspace{-5mm}
\end{figure}

To better understand how structured residual routing changes DiT training, we analyze the internal representation and gradient dynamics of DiT-XL/2 in Figure~\ref{fig:analysis}(a)--(c).

\paragraph{Representation structure.}
Figure~\ref{fig:analysis}(a) visualizes the CKA representation similarity of the baseline DiT-XL/2.
The baseline shows broadly high similarity across many layers, indicating that representations are strongly homogenized along depth.
Such high similarity suggests that the standard residual stream tends to mix layer information into a monolithic state, making it difficult for different depths to maintain distinct functional roles.
In contrast, Figure~\ref{fig:analysis}(b) shows that our method reduces excessive feature similarity and produces a more structured representation pattern.
In particular, the anti-diagonal structure becomes more visible, suggesting stronger correspondence between symmetric early and late layers.
This is consistent with our motivation that mirror routing is not merely a heuristic skip connection, but aligns with a depth-wise organization in DiT.

\paragraph{Mirror-pair gradient similarity.}
Figure~\ref{fig:analysis}(c) further analyzes the cosine similarity between gradients of mirror layer pairs, where the $i$-th point corresponds to layer $i$ and layer $27-i$, and a higher value indicates more consistent optimization signals.
Compared with the baseline, our method substantially increases gradient alignment for early--late mirror pairs, while the middle layers show lower gradient similarity.
This suggests that stage-matched pathways coordinate fine-grained feature encoding and reconstruction, while the middle layers retain distinct roles; our method thus does not simply make all layers more similar, but reorganizes gradient flow into a more structured mirror-pair pattern.
As this alignment is partly induced by the mirror connections themselves, we present it as a characterization of the optimization behavior induced by our routing (details in Appendix~\ref{sec:app_analysis}).

\subsection{Ablation studies}
\label{sec:ablation}

\begin{table}[t]
\centering
\caption{
Ablation on cross-layer connectivity on DiT-S/2 (400K iterations).
U-ViT-style skip and AttnRes are our reimplementations in DiT under the same training protocol.
\textbf{Enc.}: whether encoder blocks route over preceding encoder representations;
\textbf{Pool}: candidate sources (encoder-side representations, or all preceding block outputs);
\textbf{Selection}: sources available to each decoder sublayer;
\textbf{Fusion}: static (a learned linear projection of the concatenated features, as in U-ViT) or adaptive (token-dependent softmax routing).
}
\label{tab:ablation}
\vspace{1mm}
\setlength{\tabcolsep}{5pt}
\renewcommand{\arraystretch}{0.95}
\scriptsize
\scalebox{1.1}{\begin{tabular}{lcllc ccccc}
\toprule
Method & Enc. & Pool & Selection & Fusion & FID$\downarrow$ & sFID$\downarrow$ & IS$\uparrow$ & Pre.$\uparrow$ & Rec.$\uparrow$ \\
\midrule
DiT-S/2 & -- & -- & -- & Identity & 69.81 & 13.16 & 19.8 & 0.36 & 0.57 \\
\midrule
U-ViT-style skip & $\times$ & Encoder & Mirror & Static & 67.90 & \textbf{12.60} & 20.1 & 0.37 & 0.55 \\
Fused-Mirror w/o Enc. & $\times$ & Encoder & Mirror & Adaptive & 64.83 & 12.89 & 21.6 & 0.38 & 0.57 \\
\midrule
AttnRes & $\checkmark$ & All history & All & Adaptive & 62.87 & 12.87 & 22.6 & \textbf{0.39} & \textbf{0.58} \\
\midrule
Fused-All & $\checkmark$ & Encoder & All & Adaptive & 63.48 & 12.68 & 22.4 & \textbf{0.39} & 0.57 \\
Fused-First & $\checkmark$ & Encoder & First & Adaptive & 63.51 & 12.86 & 22.5 & 0.38 & 0.57 \\
Fused-Last & $\checkmark$ & Encoder & Last & Adaptive & 66.05 & 12.73 & 21.3 & 0.37 & 0.56 \\
\textbf{Fused-Mirror (Ours)} & $\checkmark$ & Encoder & Mirror & Adaptive & \textbf{62.48} & 12.91 & \textbf{22.8} & \textbf{0.39} & \textbf{0.58} \\
\bottomrule
\end{tabular}}
\vspace{-3mm}
\end{table}

Table~\ref{tab:ablation} studies the impact of residual formulation and source connectivity.
Unless otherwise specified, all ablation experiments are conducted on DiT-S/2.

\paragraph{Fusion: adaptive vs.\ static.}
As discussed in Section~\ref{sec:discussion}, mirror routing shares the stage pairing of U-ViT but differs in how the paired features are fused.
To verify this, we reproduce a U-ViT-style skip under our training protocol and compare it with Fused-Mirror w/o Enc., which connects the same mirror pairs on the same standard encoder but fuses them with content-dependent routing instead of a static projection.
Despite about $77\times$ more parameters, the static skip stays close to the baseline throughout training (Table~\ref{tab:traj}; $67.90$ vs.\ $69.81$ FID at $400$K), whereas adaptive fusion over the same pairs reaches $64.83$.
This advantage arises because our routing weights are computed from the token representations and can adapt to different inputs and denoising stages, making the fusion far more flexible than a fixed projection.
Adding encoder-internal routing further improves FID to $62.48$.

\paragraph{Source pool: all history vs.\ encoder.}
AttnRes is our implementation of Attention Residuals~\citep{chen2026attnres} under the same training and evaluation protocol, which uses the same source-scoring operator as ours but routes densely over all preceding block outputs, incurring a much larger activation overhead (Table~\ref{tab:cost} in Appendix~\ref{sec:app_results}).
Of the two differences from AttnRes discussed in Section~\ref{sec:discussion}, Fused-All isolates the first: it keeps dense routing but restricts the source pool to encoder-side representations.
It attains a similar FID ($63.48$ vs.\ $62.87$) with a smaller source pool, suggesting that reusing decoder-layer outputs as sources is unnecessary for image generation. We next examine the effect of the second difference, stage-matched source selection.

\paragraph{Source selection: mirror vs.\ all, first, and last.}
With the encoder-side pool fixed, the Fused variants differ only in the sources available to each decoder sublayer.
Mirror routing achieves the best FID and IS among them.
Although all routing is nominally more flexible and already prefers mirrored/symmetric layer pairs (Figure~\ref{fig:attention}, left), it has to discover this correspondence among many candidates; mirror routing provides it directly as a structural prior, reducing routing ambiguity and simplifying optimization, whereas first or last routing lacks stage correspondence.
Mirror routing thus balances flexibility and structure, avoiding the redundancy of dense routing while preserving stage-matched cross-depth interaction.

\section{Conclusion}

In this work, we revisit residual connectivity in Diffusion Transformers as an adaptive source routing problem.
We propose a structured residual routing framework that collects differentiable encoder-side representations and routes them to stage-matched decoder layers for residual reconstruction.
Our analysis shows that the proposed mirror routing reduces excessive feature homogenization and introduces clearer depth-wise representation and gradient structure.
Experiments on ImageNet demonstrate consistent improvements across DiT scales and REPA-based fine-tuning with negligible parameter overhead, while matching dense Attention Residuals at a much lower activation cost.
These results suggest that better residual connectivity can improve both convergence and model expressiveness.
We hope our findings encourage further exploration of structured information and gradient routing in scalable generative models.

\bibliography{main}
\bibliographystyle{iclr2027_conference}

\newpage
\appendix
\section{Additional experimental details}
\label{sec:app_details}

We conduct experiments on ImageNet $256\times256$ class-conditional generation.
For a fair comparison, DiT-based experiments follow the original DiT training protocol~\citep{peebles2023scalable}, while REPA-based fine-tuning starts from the released REPA checkpoint and only adds our structured residual routing modules, using the denoising objective without the REPA projection loss; the continued fine-tuning baseline uses the identical setting without routing.
Unless otherwise specified, we use a learning rate of $1\times10^{-4}$
and a total batch size of $256$.
All results are reported without classifier-free guidance unless explicitly stated.

All experiments are conducted using NVIDIA H200 GPUs.
The controlled DiT-S/2, DiT-B/2, and DiT-XL/2 experiments follow the training budgets reported in the main paper, and the REPA-based experiment fine-tunes the pretrained model with our routing modules for an additional $0.28$M and $0.35$M steps.

\section{Additional quantitative results}
\label{sec:app_results}

Table~\ref{tab:cost} compares the routing overhead of the variants discussed in Section~\ref{sec:main}, and Table~\ref{tab:traj} reports the training trajectories of DiT-S/2 underlying the ablation in Section~\ref{sec:ablation}.

\begin{table}[ht]
\centering
\caption{
Routing overhead on DiT-S/2.
Sources: candidate sources per decoder sublayer; FLOPs are relative to the DiT-S/2 baseline.
Routing act.: activation of the decoder-side source stacks (batch size 64 per GPU, 16-bit); 
encoder-side routing (absent in the U-ViT-style skip and Fused-Mirror w/o Enc.) is excluded.
}
\label{tab:cost}
\vspace{1mm}
\setlength{\tabcolsep}{6pt}
\footnotesize
\begin{tabular}{lccccc}
\toprule
Method & Sources & Params & FLOPs & Routing act. & FID$\downarrow$ \\
\midrule
U-ViT-style skip & $\boldsymbol{O(1)}$ & +1.476M & +6.9\% & -- & 67.90 \\
AttnRes & $O(L)$ & \textbf{+0.019M} & +0.7\% & 1.51 GB & 62.87 \\
Fused-All & $O(L)$ & \textbf{+0.019M} & +0.5\% & 1.20 GB & 63.48 \\
\textbf{Fused-Mirror (Ours)} & $\boldsymbol{O(1)}$ & \textbf{+0.019M} & \textbf{+0.3\%} & \textbf{289 MB} & 62.48 \\
\bottomrule
\end{tabular}
\end{table}

\begin{table}[ht]
\centering
\caption{
FID of DiT-S/2 during training (ImageNet $256\times256$, w/o CFG).
The U-ViT-style skip stays close to the standard DiT throughout training, whereas our routing is better at every checkpoint.
}
\label{tab:traj}
\vspace{1mm}
\setlength{\tabcolsep}{6pt}
\small
\begin{tabular}{lccccc}
\toprule
Model & 100K & 200K & 300K & 400K & 500K \\
\midrule
DiT-S/2 & 95.80 & 79.39 & 72.55 & 69.81 & 66.20 \\
+ U-ViT-style skip & 97.07 & 79.65 & 72.10 & 67.90 & 65.45 \\
+ Ours & 89.63 & 73.11 & 66.72 & 62.48 & 60.43 \\
\bottomrule
\end{tabular}
\end{table}

\section{Additional analysis}
\label{sec:app_analysis}

\paragraph{Mirror-pair gradient similarity.}
For Figure~\ref{fig:analysis}(c), we compute, for each mirror pair of DiT-XL/2 at $400$K iterations, the cosine similarity between the gradients with respect to the outputs of layer $i$ and layer $27-i$ under a fixed probe loss (the squared output norm on four random latents at $t{=}500$).
Gradient similarity measures whether two layers receive similar optimization signals.
A higher value indicates that the two layers are being optimized toward more consistent objectives, while a lower value indicates functional differentiation.
Since early and late layers are both closely related to dense visual details and high-frequency reconstruction, the increased alignment of early--late mirror pairs suggests that stage-matched pathways coordinate fine-grained feature encoding and reconstruction.
Meanwhile, the middle layers show lower gradient similarity, indicating that they remain functionally differentiated and can focus more on semantic abstraction and transformation.
The increase near the center is also expected, since these layers are closer in depth and naturally operate at more similar functional stages.

\section{Additional visualization results}

\begin{figure}[h!]
    \centering
    \includegraphics[width=0.89\linewidth]{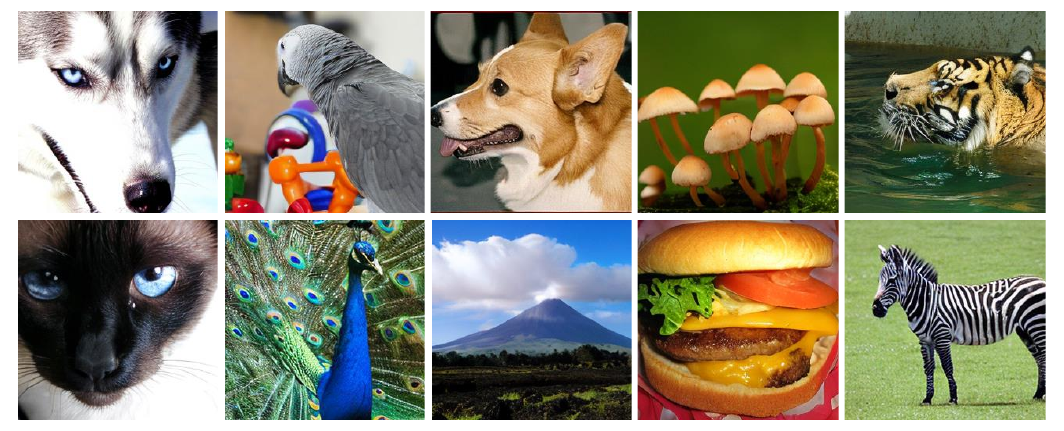}
    \caption{Additional qualitative results on ImageNet $256\times256$.
Samples are generated by the REPA-XL/2 model fine-tuned with our structured residual routing.
We use classifier-free guidance with guidance scale $w=4.0$.}
    \label{fig:append}
\end{figure}

We present additional qualitative samples generated by our method on ImageNet $256\times256$. We use the REPA-XL/2 model fine-tuned with the proposed structured residual routing and apply classifier-free guidance with guidance scale $w=4.0$, as shown in Figure~\ref{fig:append}. These results complement the main quantitative comparisons and show that our method produces visually faithful and high-quality samples across diverse ImageNet categories.

\section{Limitations}

Although the proposed structured residual routing consistently improves DiT-S/2, DiT-B/2, DiT-XL/2, and REPA-based fine-tuning, broader validation on larger text-to-image and text-to-video models remains future work.
These models usually involve more complex conditioning mechanisms, larger-scale datasets, and different training recipes, which may affect the optimal routing structure.
It is also unclear whether the current mirror routing strategy is always the best choice when stronger multimodal conditioning or temporal dependencies are introduced.
Future work could explore how structured residual routing generalizes to large-scale multimodal generation and video diffusion transformers.


\section{Broader impact}

This work aims to improve the training efficiency and generation quality of diffusion transformer models by designing better residual connectivity. 
The positive impact of this research includes enhancing high-quality generative modeling of scalable visual generation systems. 

At the same time, improvements in image generation models may also increase the risk of misuse, such as generating misleading or synthetic visual content. 
Our work does not introduce a new dataset or release a high-risk pretrained generative model, but it contributes to the broader progress of generative modeling. 
We encourage responsible use of generative models, including appropriate disclosure of synthetic content, careful dataset curation, and safeguards when deploying such systems in real-world applications.




\end{document}